\documentclass[letterpaper]{article} 
\usepackage[submission]{aaai23}  
\usepackage{times}  
\usepackage{helvet}  
\usepackage{courier}  
\usepackage[hyphens]{url}  
\usepackage{graphicx} 
\usepackage{natbib}  
\usepackage{caption} 
\usepackage{algorithm}
\usepackage{algorithmic}

\usepackage{multirow} 
\usepackage{makecell}
\usepackage{amsmath}
\usepackage{amsfonts}
\usepackage{hyperref}

\usepackage{soul}
\usepackage{color}
\usepackage{newfloat}
\usepackage{listings}
\DeclareCaptionStyle{ruled}{labelfont=normalfont,labelsep=colon,strut=off} 
\floatstyle{ruled}
\newfloat{listing}{tb}{lst}{}
\floatname{listing}{Listing}
\title{CycleTrans: Learning Neutral yet Discriminative Features for \\ Visible-Infrared Person Re-Identification}
\author{
Qiong Wu$^{1,2}$, Jiaer Xia$^{2}$, Pingyang Dai$^{2}$\thanks{Corresponding Author.}, Yiyi Zhou$^{2,4}$, Yongjian Wu$^{3}$, Rongrong Ji$^{1,2,4}$\\
$^1$Institute of Artificial Intelligence, Xiamen University. \\
$^2$Media Analytics and Computing Lab, Department of Artificial Intelligence, \\
School of Informatics, Xiamen University, 361005, China.\\
$^3$Tencent Youtu Lab.
$^4$Peng Cheng Laboratory, Shenzhen, China.\\
{\tt\small \{qiong, xiajiaer\}@stu.xmu.edu.cn, \{pydai, zhouyiyi\}@xmu.edu.cn,\\
\tt\small \{littlekenwu@tencent.com, rrji@xmu.edu.cn}
}
\usepackage{bibentry}

\begin{document}
\maketitle

\begin{abstract}
Visible-infrared person re-identification (VI-ReID) is a task of matching the same individuals across the visible and infrared modalities.
Its main challenge lies in the modality gap caused by cameras operating on different spectra.
Existing VI-ReID methods mainly focus on learning general features across modalities, often at the expense of feature discriminability.
To address this issue, we present a novel cycle-construction-based network for neutral yet discriminative feature learning, termed \emph{CycleTrans}.
Specifically, CycleTrans uses a lightweight \emph{Knowledge Capturing Module} (KCM) to capture rich semantics from the modality-relevant feature maps according to pseudo queries.
Afterwards, a \emph{Discrepancy Modeling Module} (DMM) is deployed to transform these features into neutral ones according to the modality-irrelevant prototypes.
To ensure feature discriminability, another two KCMs are further deployed for feature cycle constructions. 
With cycle construction, our method can learn effective neutral features for visible and infrared images while preserving their salient semantics.
Extensive experiments on SYSU-MM01 and RegDB datasets validate the merits of CycleTrans against a flurry of state-of-the-art methods, $+4.57\%$ on rank-1 in SYSU-MM01 and $+2.2\%$ on rank-1 in RegDB.
\end{abstract}

\section{Introduction}

Visible-infrared person re-identification (VI-ReID) aims at matching visible and infrared images of pedestrians with the same identity, which are captured by cameras operating on different spectra.
As more and more infrared cameras are deployed in real-world scenarios with dark environments, the research of VI-ReID has attracted increasing attention from both academia and industry \cite{conf/iccv/WuZYGL17, conf/ijcai/DaiJWWH18, conf/mm/YeLL19, conf/cvpr/LuWLZLCY20, conf/iccv/FuH0S0H21, journals/pami/YeSLXSH22}. 
In addition to the intrinsic challenges of traditional Re-ID tasks, such as the variations of viewpoints, illumination, and body poses, VI-ReID also suffers from the obvious appearance difference between pedestrian images of different modalities \cite{conf/cvpr/0004GLL019, journals/corr/abs-2203-01675}.

\begin{figure}[t]
\centering
\includegraphics[width=1.0\columnwidth]{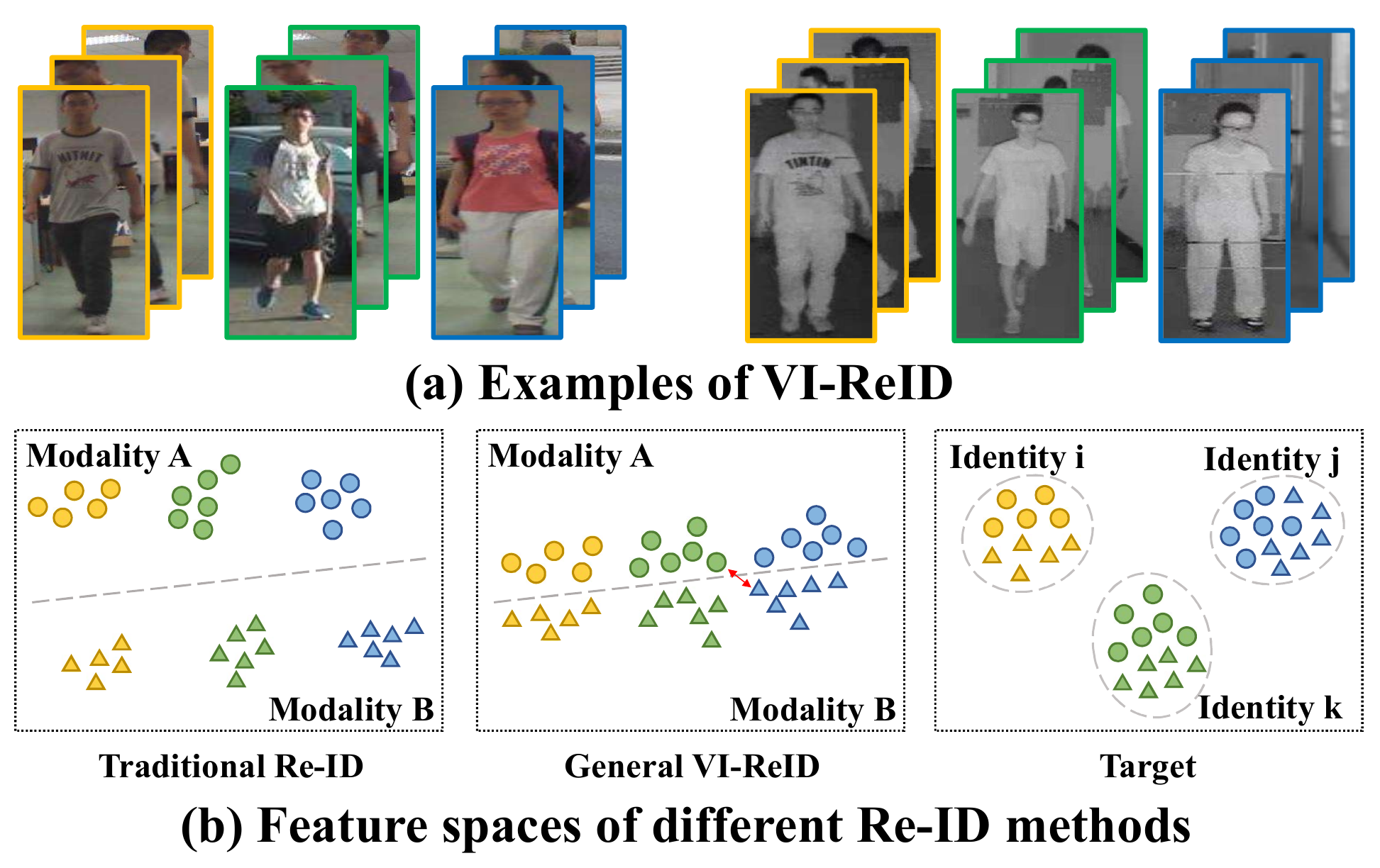}
\caption{\small
Illustrations of the examples of visible-infrared person re-identification (VI-ReID) and the feature spaces of different Re-ID methods.
(a) In VI-ReID, pedestrians with the same identity exhibit notable appearance differences between visible and infrared images, which is often termed \emph{modality gap}.
(b) In VI-ReID, traditional methods (left) often fail to match pedestrians across modalities, while the neutral features learned by existing VI-ReID methods (middle) are still not discriminative enough for identification. 
}
\label{fig1:Motivation}
\end{figure}

This issue is also coined as the modality gap \cite{journals/ijcv/WuZGL20, conf/iccv/HaoZYS21}, as illustrated in Fig.~\ref{fig1:Motivation}-(a).
Specifically, under different types of cameras, the pedestrian will exhibit notable differences in visual characteristics, \emph{e.g.}, the color and texture of clothes. 
And this gap will be further reflected in the features extracted by deep neural networks, as shown in Fig.~\ref{fig1:Motivation}-(b).  
In this case, the traditional Re-ID methods \cite{conf/eccv/SunZYTW18, conf/eccv/ZhongZLY18, conf/cvpr/0004GLL019, conf/iccv/He0WW0021}, which identify pedestrians mainly based on the appearance, often fail to accomplish this task.

In recent years, a bunch of methods have been proposed for VI-ReID and achieved remarkable progress \cite{conf/cvpr/WangWZCS19, conf/iccv/WangZ0LYH19, conf/aaai/WangZYCCLH20, conf/aaai/LiWHG20, conf/ijcai/DaiJWWH18, conf/mm/PuC0BL20, journals/corr/abs-2007-09314}. 
One main solution is to generate the representation of a missing or middle modality via generative networks \cite{conf/cvpr/WangWZCS19, conf/iccv/WangZ0LYH19, conf/aaai/WangZYCCLH20, conf/aaai/LiWHG20}. 
The synthesized images or feature maps are then regarded as the real ones to align two modalities, which also benefit the generalization of networks as additional samples.
Unfortunately, there may be distribution differences that make the optimization go awry and the performance is also limited by the quality of generated samples \cite{conf/cvpr/WangWZCS19, conf/aaai/LiWHG20}.
The other one is to learn general features for visible and infrared images via building a common embedding space \cite{conf/ijcai/DaiJWWH18, conf/mm/PuC0BL20, journals/corr/abs-2007-09314, zhou2019plenty}.
These methods avoid the noise in sample generation and do not require additional computation.
However, aligning the distributions of two modalities without complementary information tends to lose the salient semantics of each modality during transformation, resulting in a reduction in discriminative information, as shown in Fig. \ref{fig1:Motivation}-(b).
%

%

\begin{figure*}[t]
\centering
\includegraphics[width=1.0\textwidth]{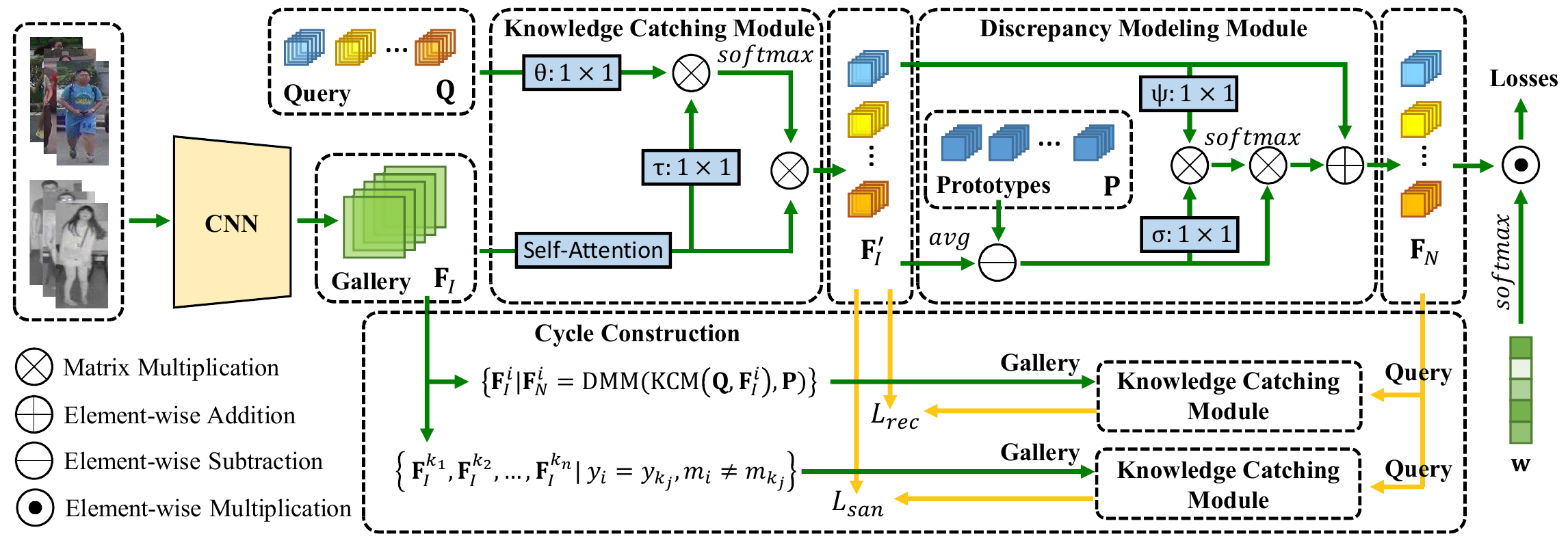}
\caption{ \small
The overview of the proposed CycleTrans.
Given an image of arbitrary modality, CycleTrans first use the proposed Knowledge Capturing Module (KCM) to gather salient yet task-related semantics from convolution feature maps based on the modality-relevant pseudo queries.
Afterwards, the Discrepancy Modeling Module (DMM) is deployed to transform these features into neutral ones via modeling the discrepancy to modality-irrelevant prototypes.
To ensure feature discriminability, a cycle construction stage is implemented (bottom), where another two KCMs are used to transform neutral features into the original modality-relevant representations.
}
\label{fig2:Framework}
\end{figure*}

To overcome these limitations, we propose a novel cycle-construction-based network (\emph{CycleTrans}) for VI-ReID.
%
The main principle of CycleTrans is to enhance the descriptive power of transformed neutral features via semantical cycle reconstructions.
As shown in Fig.~\ref{fig2:Framework}, the proposed CycleTrans consists of three Knowledge Capturing Modules (KCMs) sharing the same parameters, and a Discrepancy Modeling Module (DMM).
Specifically, the first KCM extracts discriminative semantics from convolution feature maps according to modality-specific queries.
%
%
%
Afterwards, DMM is applied to transform these features into neutral ones for visible and infrared images, which aligns the cross-modal semantics according to modality-irrelevant prototypes.
%
%
To ensure discriminability, another two KCMs are further applied to reconstruct the modality-relevant features learned before. 
Through this cycle construction process, the proposed method can well model general features across modalities while preserving salient semantics for fine-grained pedestrian identification.

To validate the proposed CycleTrans, we conduct extensive experiments on SYSU-MM01 \cite{conf/iccv/WuZYGL17} and RegDB \cite{journals/sensors/NguyenHKP17} benchmarks. 
The experimental results not only show its obvious performance gains over the SOTA methods, \emph{e.g.} $+4.57\%$ Rank-1 than SMCL \cite{conf/iccv/Wei00021} in SYSU-MM01 and $+2.2\%$ Rank-1 than FMCNet~\cite{Zhang_2022_CVPR} in RegDB, respectively, but also greatly confirm its effectiveness towards the modality gap. 

Overall, our main contributions are three-fold:
\begin{itemize}
    \item We propose a novel cycle-construction-based network for VI-ReID, termed CycleTrans.
     CycleTrans applies shared prototypes as transferring targets to mitigate the modality gap, and adopts the cycle construction to enhance feature discriminability. 
    \item To mine semantic knowledge and alleviate the modality gap, two novel modules are proposed, namely Knowledge Capturing Module (KCM) and Discrepancy Modeling Module (DMM), which can help the model learn discriminative yet neural features.
    \item The proposed CycleTrans achieves new SOTA performance on multiple benchmark datasets, \emph{e.g.}, $71.96\%$ on Rank-1 in SYSU-MM01 under \emph{single-shot} \emph{all-search} setting. 
    And the experimental results also well validate its effectiveness toward the modality gap. 
\end{itemize}

\section{Related Work}


Visible-infrared person re-identification (VI-ReID) is a task of matching the same individuals across the visible and infrared modalities.
It makes up for the failure of a visible camera to capture meaningful information in dark environments.
Besides the inherent challenge of varying viewpoints, illumination, and body poses~\cite{journals/ijon/ChenYXG21, conf/cvpr/WangZLZZ16, conf/iccv/BryanGZP19, conf/iccv/ZhouWHW19}, VI-ReID primarily faces the challenge of the modality gap that results in an obvious appearance difference for an individual.
%

%

Wu~\emph{et al.}~\cite{conf/iccv/WuZYGL17} first release a large-scale visible-infrared dataset termed SYSU-MM01 and propose a deep zero-padding network to accomplish a cross-modality task.
Some methods~\cite{conf/aaai/HaoWLG19, journals/tip/FengLX20, journals/tip/ZhangYWLZZ21} use a two-stream model to extract features from different modalities, which constrain the modality variations at both the feature and prediction levels.
Recently, MSO~\cite{conf/mm/GaoL0GLLL21} and  CoAL~\cite{conf/mm/WeiLHKG20} are proposed to capture intra-modality information, to improve feature discriminability.
cmGAN~\cite{conf/ijcai/DaiJWWH18} presents the first attempt of using generative adversarial methods for VI-ReID.
Inspired by cmGAN, several subsequent works also apply generative networks to VI-ReID.
AlignGAN~\cite{conf/iccv/WangZ0LYH19} and JSIA~\cite{conf/aaai/WangZYCCLH20} adopt GANs to generate the images of the missing modality to align cross-modal distributions at the pixel and feature levels.
$\rm D^2RL$~\cite{conf/cvpr/WangWZCS19} proposes to supplement the missing modality by GAN and unify images into four dimensions containing both RGB and infrared.
Meanwhile, X-modality~\cite{conf/aaai/LiWHG20}, cm-SSFT~\cite{conf/cvpr/LuWLZLCY20}, and  MSA~\cite{conf/ijcai/MiaoLSXY21} introduce an extra modality between the visible and infrared ones instead of directly transferring them to the opposite modality.
Nevertheless, the additional modality used in cm-SSFT requires additional modality information even during testing.
In contrast to the methods mentioned above, the proposed CycleTrans transforms both two modalities into a common distribution according to the modality-irrelevant prototypes and preserve discriminability via semantical cycle constructions. 




\section{Preliminary}

Let $\mathcal{D}=\{(\mathbf{x}_i, \mathbf{y}_i, \mathbf{m}_i)\}_{i=1}^{N}$ denotes the visible-infrared person re-identification (VI-ReID) dataset which has $N$ samples in total.
For each example, denoted as $(\mathbf{x}_i, \mathbf{y}_i, \mathbf{m}_i)$, the image $\mathbf{x}_i$ has a corresponding identity label $\mathbf{y}_i \in \mathcal{Y}=\{\mathbf{y}_j\}_{j=1}^{N_p}$ and a modality label $\mathbf{m}_i \in \mathcal{M}=\{v, r\}$, where $N_p$ is the number of identities. 
$v$ and $r$ denote the visible and infrared modalities, respectively.
Given a query image of a pedestrian, VI-ReID aims to match the same person in the other modality by ranking the similarity to instances in the gallery set, and its objective can be defined by:
\begin{equation}
\begin{aligned}
\operatorname*{argmin}_{\Theta}\mspace{-27mu}\sum\limits_{\substack{\mathbf{x}_i, \mathbf{x}_j, \mathbf{x}_k \\ \mathbf{y}_i = \mathbf{y}_j, \mathbf{y}_i \neq \mathbf{y}_k}} \mspace{-20mu} I\big(d_\Theta(\mathbf{x}_i, \mathbf{x}_j) > d_\Theta(\mathbf{x}_i, \mathbf{x}_k)\big)
\end{aligned}
\end{equation}
where $I(\cdot)$ is an indicator function that returns $1$ if the condition is satisfied and $0$ otherwise, and $d_\Theta(\cdot, \cdot)$ measure the distance between two features extracted by the model with parameters $\Theta$.

\section{Method}
\subsection{Overview}
The overall structure of the proposed cycle-construction-based network (\emph{CycleTrans}) is depicted in Fig.~\ref{fig2:Framework}.
Its main principle is to enhance the descriptive power of the transformed neutral features via feature cycle constructions.

Specifically, for a visible or infrared image $\mathbf{x}_i$, we first apply a convolutional backbone to extract its feature map, denoted as $\mathbf{F}_I\in \mathbb{R}^{h\times w\times d}$, where $h\times w$ denotes the resolution and $d$ is the number of feature dimension.
Afterwards, we use the proposed Knowledge Capturing Module (KCM) to mine the rich semantics from $\mathbf{F}_{I}$:
\begin{equation}
\mathbf{F}_{I}'=KCM(\mathbf{F}_I, \mathbf{Q}),
\end{equation} 
where $\mathbf{Q} \in \mathbb{R}^{k\times d}$ denotes the trainable pseudo queries of the corresponding modality. 
After the process of KCM, the obtained features $\mathbf{F}_I' \in \mathbb{R}^{k\times d}$ contains descriptive semantics for Re-ID, but it is still modality-relevant. 

To this end, we further transform $\mathbf{F}_I'$ into neutral features via a novel Discrepancy Modeling Module (DMM):
\begin{equation}
\mathbf{F}_{N}=DMM(\mathbf{P}, \mathbf{F}_I'),
\end{equation}
where $\mathbf{P}\in \mathbb{R}^{n \times d}$ are modality-irrelevant prototypes. 
Neutral features $\mathbf{F}_N \in \mathbb{R}^{k \times d}$ is further flattened to a representation vector and then used for cross-modal retrieval. 

To ensure the discriminability of the transformed $\mathbf{F}_N$ , we also use it to reconstruct the modality-relevant features $\mathbf{F}_I'$ via another two KCMs. 
To keep the model compact, the three KCMs share the same parameters.  

Overall, through this cycle-construction paradigm, the proposed CycleTrans can well capture salient semantics from each modality and learn effective neutral representations for cross-modal retrieval.  


\vspace{1mm}
\subsection{Knowledge Capturing Module}
\vspace{1mm}

Knowledge Capturing Module (KCM) is a novel and lightweight module for learning discriminative and task-related semantics from convolutional feature maps.

Concretely, given the feature map of an arbitrary modality $\mathbf{F}_I \in \mathbb{R}^{h \times w \times d}$, we first reshape it to a 2-$d$ tensor $\hat{\mathbf{F}}_I \in \mathbb{R}^{hw \times d}$. 
Then, we apply a \emph{dot-product} attention to refine the features by aggregating semantics from similar regions:
\begin{equation}
\tilde{\mathbf{F}}_I=Softmax\big( \text{norm}(\hat{\mathbf{F}}_I) \text{norm}(\hat{\mathbf{F}}_I)^T \big)\hat{\mathbf{F}}_I,
\end{equation}
where $\text{norm}(\cdot)$ denotes $l$-2 normalization.
The obtained feature map $\tilde{\mathbf{F}}_I$ mainly represents the general semantics of a given image, while the relevant ones for pedestrian identification still need to be enhanced.
%

In this case, we implement a cross-attention operation to mine task-related semantics based on the learn-able pseudo queries $\mathbf{Q}\in \mathbf{R}^{k\times d}$:
\begin{equation}
\begin{aligned}
\mathbf{F}'_I=Softmax(\frac{\mathbf{Q}\mathbf{W_\theta}(\Tilde{\mathbf{F}}_I\mathbf{W_\tau})^T}{\sqrt{c}})\Tilde{\mathbf{F}}_I,
\end{aligned}
\label{equation_cross_attention}
\end{equation}
where $\mathbf{W}_\theta$ and $\mathbf{W}_\tau$ are weight matrices.
Since the pseudo queries $\mathbf{Q}$ are highly task-related, they can well help the model mine useful semantics for VI-ReID via Eq.\ref{equation_cross_attention}, resulting in more discriminative features.

Notably, in our CycleTrans, KCM first serves to extract modality-relevant features for VI-ReID.
During the cycle construction, KCM is used as a module to recover modality-specific information based on the neutral features, which can be achieved by placing different feature maps as the gallery.

\vspace{2mm}
\subsection{Discrepancy Modeling Module}
\vspace{2mm}

Discrepancy Modeling Module (DMM) acts to mitigate the modality gap of VI-ReID.
Instead of directly embedding the modality-relevant features into a common semantic space, DMM learns the neutral features via aggregating information from a set of modality-irrelevant prototypes.

Concretely, given the discriminative features learned by KCM, denoted as $\mathbf{F}'_I$, we first calculate their discrepancy to the trainable prototypes $\mathbf{P} \in \mathbb{R}^{n \times d}$, where $n$ is the number of prototypes:
\begin{equation}
\begin{aligned}
\mathbf{P}'= \mathbf{P}-\hat{\mathbf{f}_I},
\end{aligned}
\label{equation_calculate_discrepancy}
\end{equation}
where $\mathbf{P}'\in \mathbb{R}^{n\times d}$ refers to the obtained discrepancy tensor and $\hat{\mathbf{f}}_I$ denotes the averaged feature of $\mathbf{F}_I'$.
Afterwards, the neutral features $\mathbf{F}_N \in \mathbb{R}^{k\times d}$ are obtained via a residual connection and a cross-attention:
\begin{equation}
\begin{aligned}
\mathbf{F}_N &= \mathbf{F}_I' + \mathbf{A}\mathbf{P}', \\
\text{where} \ \mathbf{A} &= Softmax(\frac{\mathbf{F}'_I\mathbf{W_\psi}(\mathbf{P}\mathbf{W_\sigma})^T}{\sqrt{c}}).
\end{aligned}
\label{equation_generating_mapping_vectors}
\end{equation}
Here, the attention weights $\mathbf{A}\in \mathbb{R}^{k \times n}$ are also the weighted adjacent matrix between $\mathbf{F}_I'$ and $\mathbf{P}$. 
$\mathbf{W_\psi}$ and $\mathbf{W_\sigma}$ are weight matrices.
%
%
$\mathbf{A}$ can help to reformulate semantics in a general space according to the modality-relevant features $\mathbf{F}'_I$.

Note that the sum of each row in $\mathbf{A}$ equals to 1, and $\mathbf{P}'=\mathbf{P}-\hat{\mathbf{f}}_I$. 
Thus, Eq.\ref{equation_generating_mapping_vectors} can be rewritten as:
\begin{equation}
\mathbf{F}_N = \mathbf{F}'_I + \mathbf{A} (\mathbf{P}-\hat{\mathbf{f}}_I),
\label{equation_summary_KCM_1}
\end{equation}
\begin{equation}
\mathbf{F}_N = (\mathbf{F}'_I-\hat{\mathbf{f}}_I)+ \mathbf{A} \mathbf{P}.
\label{equation_summary_KCM_2}
\end{equation}
Considering $\hat{\mathbf{f}}_I$ is the averaged vector of $\mathbf{F}'_I$, the term of $(\mathbf{F}_I'-\hat{\mathbf{f}}_I)$ in Eq.\ref{equation_summary_KCM_2} will result in an informative sparse tensor.
In this case, $\mathbf{F}_N$ is mainly composed of the newly aggregated prototype features, \emph{i.e.}, $\mathbf{AP}$, thereby achieving the alignment of cross-modality distributions.
%
%

To enhance the neutral features, we also place trainable weights to adaptively adjust the contribution of each pattern in $\mathbf{F}_N$, which is achieved by:
\begin{equation}
\begin{aligned}
\mathbf{F}^{(i)}_{N} = \frac{e^{w_i}}{\sum^k_{j=1}e^{w_j}}\mathbf{F}^{(i)}_{N},
\end{aligned}
\label{equation_final_neutral}
\end{equation}
where $w_j$ is the weight for $j^{th}$ pattern of neutral feature.

From Eq.\ref{equation_summary_KCM_1}-\ref{equation_summary_KCM_2}, we can see that $\mathbf{F}_N$ contains a certain amount of modality-relevant information form $\mathbf{F}'_I- \hat{\mathbf{f}}_I$, but it is still hard to ensure that they are discriminative enough for VI-ReID.
In this case, we further implement \emph{Cycle Constructions} to enhance their descriptive power.

\subsection{Cycle Construction}
\vspace{1mm}

The main assumption of Cycle Construction is that if the learned neutral features can recover modality-relevant information well, it suggests that they are capable of both cross-modality alignment and prominent feature discrimination.

Specifically, the proposed cycle construction consists of two processes, which transform the neutral features into visible and infrared ones, respectively.
Taking a visible image for example, of which feature maps and the refined ones are denoted as $\mathbf{F}_I^v$ and $\mathbf{F}_I^{v'}$, we apply the proposed KCM to reconstruct its modality-relevant features:
\begin{equation}
\mathbf{F}_{Re}^{v}=KCM(\mathbf{F}_I^{v}, \mathbf{F}_N),
\end{equation}
where $\mathbf{F}_{Re}^{v}$ is the recovered features and $\mathbf{F}_N$ acts the role of pseudo queries described in Eq.\ref{equation_cross_attention}.
During training, we will minimize the $l$-1 distance between the recover features $\mathbf{F}_{Re}^{v}$ and the modality-relevant ones $\mathbf{F}_I^{v'}$. 

In the other stream, CycleTrans project the neutral features to the other modality with another KCM, \emph{i.e.}, the infrared one here, defined by:
\begin{equation}
\begin{aligned}
\mathbf{F}_{Re}^{r}&=KCM([\mathbf{F}^{k_1}_{I},  \mathbf{F}^{k_2}_{I},..., \mathbf{F}^{k_h}_{I}], \mathbf{F}_{N}), \\
\text{where} \ \mathbf{y}_i &= \mathbf{y}_{k_j}, \mathbf{m}_i \neq \mathbf{m}_{k_j}, j=1,2,...,h,
\end{aligned}
\label{equation_cross_construct}
\end{equation}
where $\mathbf{F}_{Re}^{r}$ denotes the recovered infrared features and $[\mathbf{F}^{k_1}_{I},  \mathbf{F}^{k_2}_{I},..., \mathbf{F}^{k_h}_{I}]$ denotes $h$ feature maps that have the same identity but from the other modality.
%

In Eq.\ref{equation_cross_construct}, the neutral features are regarded as the pseudo queries for KCM to aggregate semantics from all feature maps that may provide valuable information.
It can help to rule out the interference factors that may cause appearance discrepancy between samples for more accurate reconstruction, \emph{e.g.}, viewpoints, body poses, and obstructions.
%

%

To ensure the reconstruction, we also minimize the semantic distance between two generated features, \emph{i.e.}, $\mathbf{F}'_I$ and $\mathbf{F}_{Re}^{r}$.
This objective is also beneficial for alleviating the modality gap.
For an infrared image, the process of cycle construction is the same. 
To maintain the compactness of CycleTrans, we share the parameters of the three KCMs used.

%

\subsection{Optimization}

During training, we apply the following objectives to optimize CycleTrans.

\textbf{Cross-entropy loss.} As the core objective of VI-ReID, \emph{cross-entropy loss} is used to learn the identities of samples with classifier ${\rm C}(\cdot)$ under the supervision of the label $\mathbf{y}_i$:
\begin{equation}
\begin{aligned}
\mathcal{L}_{id}=-\frac{1}{B}\sum_{i=1}^{B}\log&{P(\mathbf{y}_i|{\rm C}(\mathbf{F}^{(i)}_{N}))},
\end{aligned}
\label{equation_identity_learning}
\end{equation}
where ${\rm C}(\mathbf{F}^{(i)}_{N})$ is the predicted identity based on $\mathbf{F}^{(i)}_{N}$.

\textbf{Metric loss.} To semantically separate the obtained neutral features, we apply a \emph{Metric loss} to CycleTrans: 
\begin{equation}
\begin{aligned}
\mathcal{L}_{me}=\frac{1}{B^2}\sum_{i=1}^B & \sum_{j=1, \mathbf{y}_i \neq \mathbf{y}_j}^B [ \rho - d(\mathbf{F}^{(i)}_{N},\mathbf{F}^{(j)}_{N}) \\
&+ d(\mathbf{F}^{(i)}_{N}, \mathbf{C}^{(i)}) + d(\mathbf{F}^{(j)}_{N}, \mathbf{C}^{(j)})]_+,
\end{aligned}
\label{loss_triplet}
\end{equation}
where $[\cdot]_+$ represents $\max\{\cdot,0\}$, $B$ denotes the batch size. 
$d(\cdot)$ is the distance function, which is $l$-2 here. 
$\mathbf{C}^{(i)}$ denotes the class center of $i^{th}$ example, which is calculated in each batch, and $\rho$ is the least margin between two classes.
Via Eq.\ref{loss_triplet}, CycleTrans can well separate the neural features of different identities and minimize the distance between the example and its multi-modality anchor, \emph{i.e.}, the class center $\mathbf{C}$. 
In this case, it is much easier to obtain the general and cross-modality representation, which is critical in VI-ReID.

\textbf{Separation loss.} To help the model learn neutral features with more diverse patterns, we define the following regularization term:
\begin{equation}
\begin{aligned}
\mathcal{L}_{sep} = \frac{1}{k^2}\sum^{k-2}_{i=1}\sum^{k-1}_{i=j+1}\frac{\mathbf{F}^{(i)}_{N}\mathbf{F}^{(j)}_{N}}{|\mathbf{F}^{(i)}_{N}|_2 |\mathbf{F}^{(j)}_{N}|_2},
\end{aligned}
\label{equation_sep}
\end{equation}
where $\mathbf{F}^{(i)}_{N}$ is the $i^{th}$ pattern of the neutral feature.
%
%
Note that, the last pattern of neutral features $\mathbf{F}_{N}$ is not involved in the $\mathcal{L}_{sep}$, which plays the role of global representation.

%
%
\textbf{Modality Fusion Loss.} We also apply the \emph{Multi-Kernel Maximum Mean Discrepancy} (MMD) \cite{journals/jmlr/GrettonBRSS12} with Gaussian kernel to alleviate the potential modality gap in neutral features:
\begin{equation}
\begin{aligned}
\mathcal{L}_{\rm MMD} = || \mathrm{E}_v[\phi({\mathbf{F}^{v}_{N}})] - \mathrm{E}_r[\phi(\mathbf{F}^{r}_{N})] ||^2_{\mathcal{H}_k}
\end{aligned}
\label{equation_MMD}
\end{equation}
where $\mathbf{\phi(\cdot)}$ is an implicit feature mapping function and $\mathcal{H}_k$ represents the \emph{Reproducing Kernel Hilbert Space} (RKHS).
${\mathbf{F}^{v}_{N}}$ and ${\mathbf{F}^{r}_{N}}$ denote the neutral features of visible and infrared images, respectively.
Eq.~\ref{equation_MMD} can ensure the consistency between the neural features of different modalities.


\textbf{Reconstruction loss.} To  ensure the discriminability of neutral features and the quality of reconstructions, we propose a distance-based reconstruction loss:
\begin{equation}
\begin{aligned}
\mathcal{L}_{rec}=|\mathbf{F}^{v}_{Re} - \mathbf{F}^{v'}_{I}|_1,
\end{aligned}
\label{equation_rec_loss}
\end{equation}
the $|\cdot|_1$ represents the $l$-1 distance. 
By decreasing the distance between reconstructed features $\mathbf{F}^{v}_{Re}$ and modality-relevant features $\mathbf{F}^{v'}_{I}$, we can keep semantic consistency during transformation.

\textbf{Alignment loss.} We also introduce an \emph{Alignment loss} to ensure the quality of recovered cross-modality features, which is defined by
\begin{equation}
\begin{aligned}
\mathcal{L}_{aln}=|\mathbf{F}^{r}_{Re} - \mathbf{F}^{v'}_{I}|_2,
\end{aligned}
\label{equation_ali_loss}
\end{equation}
where $|\cdot|$ denote the $l$-2 distance.
Eq.\ref{equation_ali_loss} can also serve to reduce the gap between visible and infrared images by aligning two types of features. 

Notably, in Eq.\ref{equation_rec_loss} and Eq.\ref{equation_ali_loss}, we use the reconstruction of visible features as an example. During training, these loss terms are also applied to infrared images.

In summary, the overall objective function of the proposed CycleTrans is defined by:
\begin{small}
\begin{equation}
\begin{aligned}
\mathcal{L}=\mathcal{L}_{id} + \mathcal{L}_{me} + \lambda_1\mathcal{L}_{sep} + \lambda_2\mathcal{L}_{\rm MMD} +  \lambda_3\mathcal{L}_{rec} + \lambda_4\mathcal{L}_{aln},
\end{aligned}
\label{equation_identity_learning}
\end{equation}
\end{small}

\noindent where $\lambda_1$, $\lambda_2$, $\lambda_3$ and $\lambda_4$ are hype-parameters.
They are set mainly based on our empirical knowledge and put the cross-entropy loss as the center.
Specifically, $\lambda_3$ and $\lambda_4$ are directly set to $0.1$ and $0.1$ based on their scales of gradients. 
Only $\lambda_1$ and $\lambda_2$ will be tuned during experiments. 

\section{Experiments}



\subsection{Datasets and Metrics}

We validate the proposed CycleTrans on two widely-used VI-ReID benchmarks, namely SYSU-MM01 \cite{conf/iccv/WuZYGL17} and RegDB \cite{journals/sensors/NguyenHKP17}.

SYSU-MM01 is a large-scale dataset consisting of both indoor and outdoor images captured by four visible cameras and two near-infrared ones.
The training set contains $395$ identities with $22,258$ visible images and $11,909$ infrared ones.
The query set has $3,803$ infrared images and the gallery set shares $96$ identities.
Under the \emph{single-shot} and \emph{multi-shot} setting, there are $301$ and $3,010$ randomly sampled visible images in the gallery, respectively.
    
RegDB is a small-scale dataset with images captured by a pair of aligned cameras (one visible and one thermal).
It contains $8,240$ images of $412$ identities, each with $10$ visible and $10$ thermal images.
The dataset is randomly divided into two splits, \emph{i.e.}, images of $206$ identities for training and the rest of $206$ identities for testing.
%

For two VI-ReID datasets, the \emph{Cumulative Matching Characteristic} (\textbf{CMC}) \cite{conf/iccv/WangDSRT07}, including \textbf{Rank-1}, \textbf{Rank-10}, and \textbf{Rank-20} accuracies, and \emph{mean Average Precision} (\textbf{\emph{m}AP}) metrics are used as the evaluation metrics.
All comparisons use the same metrics. 
%
%

\begin{table}[!t]
\centering
\renewcommand\arraystretch{1.1}
\caption{\small
Ablation study in terms of CMC (\%) and \emph{m}AP (\%) on SYSU-MM01 under \emph{single-shot all-search} setting.
}
\setlength{\tabcolsep}{1.0mm}
{
\renewcommand{\multirowsetup}{\centering}
\resizebox{0.48\textwidth}{!}
{
\begin{tabular}{l||ccc|c}
\hline
\multirow{3}{*}{\ \ \ \ Method}  & \multicolumn{4}{c}{SYSU-MM01} \\
\cline{2-5} & \multicolumn{4}{c}{Single-shot All-search}  \\
\cline{2-5} & \ \ \ Rank-1 \ \ \ & \ \ \ Rank-10 \ \ \ & \ \ \ Rank-20 \ \ \ & \ \ \ \emph{m}AP \ \ \ \\
\hline
Baseline                        & 58.99 & 91.18 & 96.06 & 54.29 \\
\hline
+ KCM                           & 62.32 & 91.53 & 96.39 & 57.49 \\
+ $\mathcal{L}_{me}$            & 64.93 & 93.88 & 97.60 & 60.74 \\
+ DMM                           & 67.40 & 94.76 & 98.15 & 63.01 \\
+ $\mathcal{L}_{rec}$           & 69.76 & 95.09 & 98.23 & 65.15 \\
+ $\mathcal{L}_{aln}$ (Full)    & 71.96 & 95.55 & 98.46 & 67.24 \\
\hline
\end{tabular}
}
\label{table_ablation}

}
\end{table}

\begin{table}[!t]
\centering
\caption{\small
The impact of different alternatives of DMM on SYSU-MM01 under \emph{single-shot all-search} setting.
}
\setlength{\tabcolsep}{1.0mm}
{
\renewcommand{\multirowsetup}{\centering}
\renewcommand\arraystretch{1.1}
\resizebox{0.48\textwidth}{!}
{
\begin{tabular}{l||ccc|c}
\hline
\multirow{3}{*}{\ \ \ \  Method}  & \multicolumn{4}{c}{SYSU-MM01} \\
\cline{2-5} & \multicolumn{4}{c}{single-shot all-search}  \\
\cline{2-5} & \ \ \ Rank-1 \ \ \ & \ \ \ Rank-10 \ \ \ & \ \ \ Rank-20 \ \ \ & \ \ \ \emph{m}AP \ \ \    \\
\hline
Baseline                & 58.99 & 91.18 & 96.06 & 54.29 \\
\hline
$\mathbf{F}_N = \mathbf{A}\mathbf{P}$                   & 68.74 & 94.80 & 98.09 & 64.32 \\
$\mathbf{F}_N = \mathbf{F}'_I + \mathbf{A}\mathbf{P}$   & 66.56 & 94.38 & 98.07 & 62.66 \\
Transformer                                             & 68.44 & 94.23 & 97.76 & 63.55 \\
\hline
DMM                     & 71.96 & 95.55 & 98.46 & 67.24 \\
\hline
\end{tabular}
}
\label{table_construction}
}
\end{table}

\subsection{Implementation details.} 

\begin{table*}[t]
\caption{
Comparison between CycleTrans and the state-of-the-art methods on SYSU-MM01.
}
\centering
\renewcommand\arraystretch{1.1}
\setlength{\tabcolsep}{0.7mm}
\resizebox{\textwidth}{!}
{
\begin{tabular}{r||ccc|c|ccc|c||ccc|c|ccc|c}
\hline
\multirow{3}{*}{Method \ \ \ \ \ \ \ }  & \multicolumn{8}{c||}{All-Search} & \multicolumn{8}{c}{Indoor-Search}    \\
\cline{2-17} & \multicolumn{4}{c|}{Single-Shot} & \multicolumn{4}{c||}{Multi-Shot} & \multicolumn{4}{c|}{Single-Shot} & \multicolumn{4}{c}{Multi-Shot} \\
\cline{2-17} & R1 & R10 & R20 & \emph{m}AP   & R1 & R10 & R20 & \emph{m}AP  & R1 & R10 & R20 & \emph{m}AP   & R1 & R10 & R20 & \emph{m}AP \\
\hline
Zero-Padding~\cite{conf/iccv/WuZYGL17}  & 14.80 & 54.12 & 71.33 & 15.95 & 19.13 & 61.40 & 78.41 & 10.89 & 20.58 & 68.38 & 85.79 & 26.92 & 24.43 & 75.86 & 91.32 & 18.86 \\
BDTR~\cite{conf/ijcai/YeWLY18} & 17.01 & 55.43 & 71.96 & 19.66 & -     & -     & -     & -     & -     & -     & -     & -     & -     & -     & -     & -     \\
D-HSME~\cite{conf/aaai/HaoWLG19} & 20.68 & 62.74 & 77.95 & 23.12 & -     & -     & -     & -     & -     & -     & -     & -     & -     & -     & -     & -     \\
cmGAN~\cite{conf/ijcai/DaiJWWH18}       & 26.97 & 67.51 & 80.56 & 27.80 & 31.49 & 72.74 & 85.01 & 22.27 & 31.63 & 77.23 & 89.18 & 42.19 & 37.00 & 80.94 & 92.11 & 32.76 \\
$\rm D^2RL$~\cite{conf/cvpr/WangWZCS19} & 28.90 & 70.60 & 82.40 & 29.20 & -     & -     & -     & -     & -     & -     & -     & -     & -     & -     & -     & -     \\
Hi-CMD~\cite{conf/cvpr/ChoiLKKK20} & 34.94 & 77.58 & - & 35.94 & -     & -     & -     & -     & -     & -     & -     & -     & -     & -     & -     & -     \\
JSIA-ReID~\cite{conf/aaai/WangZYCCLH20} & 38.10 & 80.70 & 89.90 & 36.90 & 45.10 & 85.70 & 93.80 & 29.50 & 43.80 & 86.20 & 94.20 & 52.90 & 52.70 & 91.10 & 96.40 & 42.70 \\
AlignGAN~\cite{conf/iccv/WangZ0LYH19}   & 42.40 & 85.00 & 93.70 & 40.70 & 51.50 & 89.40 & 95.70 & 33.90 & 45.90 & 87.60 & 94.40 & 54.30 & 57.10 & 92.70 & 97.40 & 45.30 \\
AGW~\cite{journals/sensors/NguyenHKP17} & 47.50 & -     & -     & 47.65 & -     & -     & -     & -     & 54.17 & -     & -     & 62.97 & -     & -     & -     & -     \\
cm-SSFT(sq)~\cite{conf/cvpr/LuWLZLCY20} & 47.70 & -     & -     & 54.10 & -     & -     & -     & -     & 57.40 & -     & -     & 59.10 & -     & -     & -     & -     \\
DFE~\cite{conf/mm/HaoWGLW19}            & 48.71 & 88.86 & 95.27 & 48.59 & 54.63 & 91.62 & 96.83 & 42.14 & 52.25 & 89.86 & 95.85 & 59.68 & 59.62 & 94.45 & 98.07 & 50.60 \\
XIV-ReID~\cite{conf/aaai/LiWHG20}       & 49.92 & 89.79 & 95.96 & 50.73 & -     & -     & -     & -     & -     & -     & -     & -     & -     & -     & -     & -     \\
CMM+CML~\cite{conf/mm/LingZLRLS20}      & 51.80 & 92.72 & 97.71 & 51.21 & 56.27 & 94.08 & \underline{98.12} & 43.39 & 54.98 & 94.38 & 99.41 & 63.70 & 60.42 & \underline{96.88} & \underline{99.50} & 53.52 \\  
SIM~\cite{conf/ijcai/JiaZLMZ20}         & 56.93 & -     & -     & 60.88 & -     & -     & -     & -     & -     & -     & -     & -     & -     & -     & -     & -     \\
CoAL~\cite{conf/mm/WeiLHKG20}           & 57.22 & 92.29 & 97.57 & 57.20 & -     & -     & -     & -     & 63.86 & 95.41 & 98.79 & 70.84 & -     & -     & -     & -     \\
MSO~\cite{conf/mm/GaoL0GLLL21}          & 58.70 & 92.06 & -     & 56.42 & 65.85 & \underline{94.37} & -     & 49.56 & 63.09 & 96.61 & -     & 70.31 & 72.06 & 91.77 & -     & 61.69 \\
DG-VAE~\cite{conf/mm/PuC0BL20}          & 59.49 & 93.77 & -     & 58.46 & -     & -     & -     & -     & -     & -     & -     & -     & -     & -     & -     & -     \\
MCLNet~\cite{conf/iccv/HaoZYS21}        & 65.40 & \underline{93.33} & \underline{97.14} & 61.98 & -     & -     & -     & -     & \underline{72.56} & \underline{96.98} & \underline{99.20} & \underline{76.58} & -     & -     & -     & -     \\
FMCNet~\cite{Zhang_2022_CVPR}           & 66.34 & -     & -     & \underline{62.51} & \underline{73.44} & -     & -     & \underline{56.06} & 68.16 & -     & -     & 74.09 & 78.86 & -     & -     & 63.82 \\
SMCL~\cite{conf/iccv/Wei00021}          & \underline{67.39} & 92.87 & 96.76 & 61.78 & 72.15 & 90.66 & 94.32 & 54.93 & 68.84 & 96.55 & 98.77 & 75.56 & \underline{79.57} & 95.33 & 98.00 & \underline{66.57} \\

\hline
CycleTrans(Ours)   & \textbf{71.96} & \textbf{95.55} & \textbf{98.46} & \textbf{67.24}
                    & \textbf{79.36} & \textbf{97.64} & \textbf{99.18} & \textbf{62.53}
                    & \textbf{82.55} & \textbf{99.58} & \textbf{99.95} & \textbf{80.46}
                    & \textbf{89.28} & \textbf{99.83} & \textbf{99.99} & \textbf{76.25} \\
\hline
\end{tabular}
}
\label{table_comparisonSYSU}
\vspace{-3mm}
\end{table*}

\begin{table}[t]
\centering{
\caption{\small Comparison with state-of-the-art methods on RegDB.}
\label{table_comparisonRegDB}
}
\resizebox{0.48\textwidth}{!}
{
\renewcommand\arraystretch{1.1}
\begin{tabular}{r||cc|cc}
\hline
\multirow{2}{*}{Method \ \ \ \ \ \ \ }  & \multicolumn{2}{c|}{Infrared to Visible} &\multicolumn{2}{c}{Visible to Infrared} \\
\cline{2-5}   & Rank-1 & \emph{m}AP & Rank-1 & \emph{m}AP  \\
\hline
Zero-Padding~\cite{conf/iccv/WuZYGL17}  & 16.7 & 17.9 & 17.8 & 18.9 \\
BDTR~\cite{conf/ijcai/YeWLY18}          & 32.7 & 31.1 & 33.5 & 31.8 \\
$\rm D^2RL$~\cite{conf/cvpr/WangWZCS19} & -    & -    & 43.4 & 44.1 \\
JSIA-ReID~\cite{conf/aaai/WangZYCCLH20} & 48.1 & 48.9 & 48.5 & 49.3 \\
D-HSME~\cite{conf/aaai/HaoWLG19}        & 50.2 & 46.2 & 50.9 & 47.0 \\
AlignGAN~\cite{conf/iccv/WangZ0LYH19}   & 56.3 & 53.4 & 57.9 & 53.6 \\
CMM+CML~\cite{conf/mm/LingZLRLS20}      & 59.8 & 60.9 & -    & -    \\
XIV-ReID~\cite{conf/aaai/LiWHG20}       & 62.3 & 60.2 & -    & -    \\
cm-SSFT(sq)~\cite{conf/cvpr/LuWLZLCY20} & 63.8 & 64.2 & 65.4 & 65.6 \\
AGW~\cite{journals/sensors/NguyenHKP17} & -    & -    & 70.0 & 66.4 \\
DFE~\cite{conf/mm/HaoWGLW19}            & 68.0 & 66.7 & 70.2 & 69.2 \\
i-CMD~\cite{conf/cvpr/ChoiLKKK20}      & -    & -    & 70.9 & 66.0 \\
DG-VAE~\cite{conf/mm/PuC0BL20}          & -    & -    & 73.0 & 71.8 \\
CoAL~\cite{conf/mm/WeiLHKG20}           & 74.1 & 69.9 & -    & -    \\
MSO~\cite{conf/mm/GaoL0GLLL21}          & 74.6 & 67.5 & 73.6 & 66.9 \\
SIM~\cite{conf/ijcai/JiaZLMZ20}         & 75.2 & 78.3 & 74.7 & 75.2 \\
MCLNet~\cite{conf/iccv/HaoZYS21}        & 75.9 & 69.5 & 80.3 & 73.1 \\
SMCL~\cite{conf/iccv/Wei00021}          & 83.1 & 78.6 & 83.9 & 79.8 \\
FMCNet~\cite{Zhang_2022_CVPR}           & \underline{89.1} & \underline{84.4} & \underline{88.4} & \underline{83.9} \\
\hline
CycleTrans(Ours)  & \textbf{91.3} & \textbf{84.88} & \textbf{90.3} & \textbf{84.9} \\
\hline
\end{tabular}
}
\vspace{-4mm}
\end{table}

For CycleTrans, we use ResNet-50~\cite{conf/cvpr/HeZRS16} as the backbone, where the stride of the last convolutional layer is set to $1$.
The classifier ${\rm C}(\cdot)$ consists of a BN neck~\cite{conf/cvpr/0004GLL019} and an FC layer without bias.
The input images are resized to $288 \times 144$ and are randomly flipped and erased \cite{conf/aaai/Zhong0KL020} with $50\%$ probability.
The $\lambda$ hyper-parameter sets are $[0.2, 1.0, 0.1, 0.1]$ for SYSU-MM01, and $[0.2, 0.8, 0.1, 0.1]$ for RegDB.
The margin $\rho$ in $\mathcal{L}_{me}$ is set to $0.5$.
The number of prototypes is set to $1024$ in our experiments.
And we apply $7$ pseudo queries for both SYSU-MM01 and RegDB to extract neutral features from both two modalities images.
During training, each mini-batch contains $64$ images of $8$ identities.
We randomly sample $4$ visible images and $4$ infrared images for each identity.
The proposed model is trained for a total of $140$ epochs and optimized by \emph{Adam} \cite{journals/corr/KingmaB14} with an initial learning rate of $3.5 \times {10}^{-4}$. 
The learning rate decays at the $40^{th}$ and $70^{th}$ epoch with a decay factor of $0.1$.

\subsection{Ablation Study}

\begin{figure*}[t]
\centering
\includegraphics[width=1.0\textwidth]{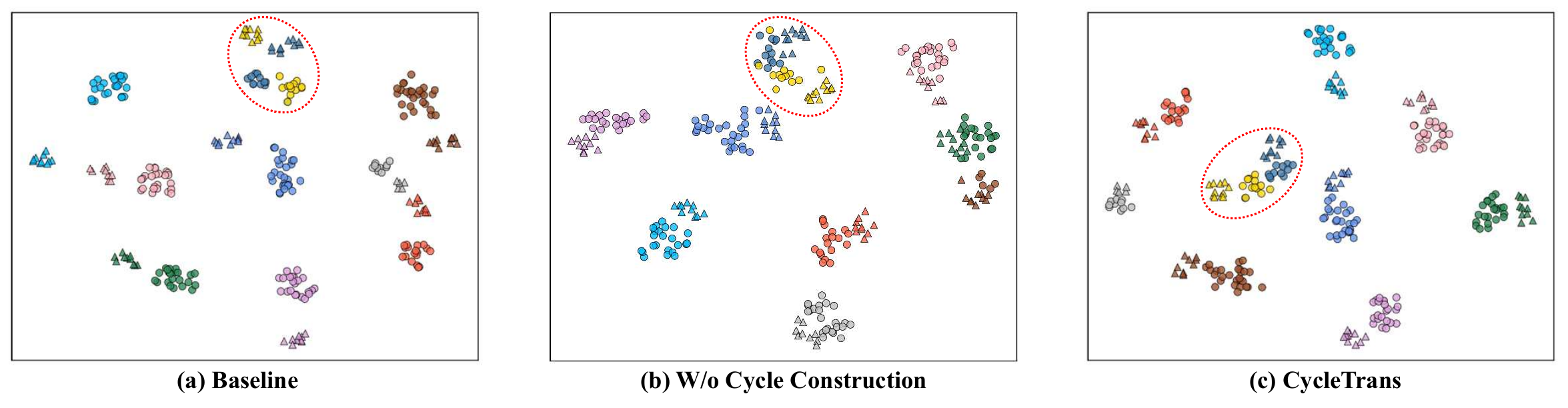}
\vspace{-6mm}
\caption{ \small
Feature visualizations.
Circles and triangles denote the features of visible and infrared images, respectively, and the colors represent different identities. 
\emph{Baseline} refers to the basic setting described in Tab. \ref{table_ablation}.
The middle plot shows the results of our CycleTrans without cycle construction.
Compared to the other two models, our CycleTrans can well cluster features of different modalities but with the same identity. 
It also exhibits more clear semantic margins between identities.  
}
\label{fig3:Distribution}
\vspace{-3mm}
\end{figure*}

We first ablate our CycleTrans on SYSU-MM01 under \emph{all-search} \emph{single-shot} setting \cite{conf/iccv/WuZYGL17}, of which results are given in Tab. \ref{table_ablation}.
Here, \emph{baseline} denotes that the model only consists of the convolution backbone and is trained merely with the cross-entropy loss $\mathcal{L}_{id}$.

Tab. \ref{table_ablation} shows the cumulative results of each design in CycleTrans. 
From this table, we can first observe that the proposed Knowledge Capturing Module (KCM) and Discrepancy Modeling Module (DMM) can significantly improve model performance, achieving $+3.33\%$ and $+2.47\%$ gains on Rank-1, respectively.
The use of cycle construction, \emph{i.e.}, $+\mathcal{L}_{rec}$ and $+\mathcal{L}_{aln}$, can also improve performance to a large extent by $+4.56\%$ on Rank-1 compared to ``$+DMM$''.
Meanwhile, we also notice that the metric loss $\mathcal{L}_{me}$ can also bring improvements on all metrics, suggesting its benefits for neutral features.
Lastly, combing all designs proposed in CycleTrans can improve the baseline by up to $+12.97\%$ Rank-1, strongly validating their effectiveness.

We also examine different alternatives of the proposed DMM module, of which results are given in Tab \ref{table_construction}.
The second block of Tab. \ref{table_construction} shows the different choices of DMM, including the one aggregating prototypes without residual connection, \emph{i.e.}, $\mathbf{F}_N = \mathbf{A}\mathbf{P}$, and the one without discrepancy modeling, \emph{i.e.}, $\mathbf{F}_N = \mathbf{F}'_I + \mathbf{A}\mathbf{P}$. 
We also use a Transformer layer \cite{conf/iclr/DosovitskiyB0WZ21} for comparison.

The first alternative only uses the aggregated prototypes as neutral features, which can strictly follow the distribution of prototype information.
However, this alternative will make the convolution backbone hard to optimize, since the image features are not directly involved in the objective functions.
Meanwhile, the lack of fine-grained image semantics from residual connection also limits its performance upper-bound.
Compared to DMM, the second alternative does not include discrepancy modeling, which leads to obvious performance degradations.
One hypothesis is that without discrepancy modeling, the obtained neutral features are still highly modality-relevant, making the model fail in cross-modal retrieval.
The use of a Transformer layer is a good choice for neutral feature transformation, which takes the modality-relevant features as queries and the prototypes as keys and values.
However, its performance is still inferior than our DMM, \emph{e.g.}, $-3.52\%$ Rank-1 and $-3.69\%$ \emph{m}AP.
Overall, these designs well confirm the effectiveness of our DMM in neutral feature learning for VI-ReID. 

\subsection{Comparison with State-of-the-art Methods}

We then compare our CycleTrans with a set of state-of-the-arts (SOTAs) on SYSU-MM01 and RegDB, of which results are given in Tab. \ref{table_comparisonSYSU} and Tab. \ref{table_comparisonRegDB}, respectively. 

\textbf{Comparisons on SYSU-MM01.}
As shown in Tab~\ref{table_comparisonSYSU}, the proposed CycleTrans outperforms existing SOTAs by large margins on SYSU-MM01.
Specifically, compared to the latest method, \emph{i.e.}, SMCL \cite{conf/iccv/Wei00021}, CycleTrans can obviously improve the performance of all metrics under \emph{All-Search} setting, \emph{e.g.}, $+4.57\%$ on Rank-1 and $+5.46\%$ on \emph{m}AP.
Under the setting of \emph{Indoor-Search}, the advantages of CycleTrans are further expanded. 
For instance, the SOTA performance on \emph{Single-shot} Rank-1 and \emph{Multi-shot} Rank-1 are improved by $+9.99\%$ and $+9.71\%$ by our method, which are indeed very significant.

\textbf{Comparisons on RegDB.}
Similar advantages of CycleTrans can be also witnessed on RegDB in Tab. \ref{table_comparisonRegDB}, which is a smaller-scale dataset. 
Under two cross-modality settings, our method achieves new SOTA performance on all metrics. 
Notably, the latest method FMCNet~\cite{Zhang_2022_CVPR} has already achieved obvious gains over previous VI-ReID methods, but our CycleTrans can further improve performance, \emph{e.g.}, $+2.2\%$ and $+1.9\%$ Rank-1 on two settings. 

Considering SYSU-MM01 and RegDB are two highly competitive benchmarks, these significant performance gains strongly validate the effectiveness of the proposed CycleTrans and our motivation about the modality gap.

\subsection{Quality Analysis}

To gain deep insight into the proposed CycleTrans, we further visualize the distributions of different features extracted by the baseline and our CycleTrans in Fig. \ref{fig3:Distribution}.
We randomly visualize samples of $10$ identities from the testing set via t-SNE \cite{van2008visualizing}.
Fig. \ref{fig3:Distribution}-(a) shows the feature distribution of the baseline. 
We can see that although these features can be mapped to different clusters, images of the same identity but different modalities are still hard to distinguish. 
For instance, the blue and yellow features of the same modalities are closely distributed in this space and hard to identify.
Fig. \ref{fig3:Distribution}-(b) shows the results of CycleTrans without Cycle Construction.
It illustrates that CycleTrans can well transform these modality-relevant features into neutral ones with the help of the proposed DMM, resulting in better clusters than Fig. \ref{fig3:Distribution}-(a).
However, due to the lack of enough feature discriminability, the cross-modality features of some identities still do not exhibit clear semantic margins, \emph{e.g.}, the yellow and blue examples.
With cycle construction, this problem is greatly alleviated, as shown in Fig.\ref{fig3:Distribution}-(c).
From this figure, we can see that our CycleTrans can learn clear margins between features of different identities.
%
Meanwhile, the better clustering result of CycleTrans than the other two methods suggests a stronger descriptive power. 
Overall, the visualization results well confirm the effectiveness of the proposed CycleTrans towards neutral yet discriminative feature learning for VI-ReID.

\section{Conclusion}

In this paper, we aim to address the modality gap in VI-ReID via learning neutral yet discriminative features. 
To approach this target, we propose a cycle-construction-based model for VI-ReID, termed CycleTrans.
Specifically, CycleTrans first use a novel Knowledge Capturing Module (KCM) to mine salient semantics from convolution feature maps based on pseudo queries. 
Afterwards, we propose a Discrepancy Modeling Module (DMM) to transform these semantics into neutral features based on the modality-irrelevant prototypes. 
To ensure the descriptive power of the neutral features, feature cycle constructions are performed via another two KCMs sharing the same parameters.
To validate our CycleTrans, we conduct extensive experiments on two highly-competitive benchmarks, namely SYSU-MM01 and RegDB. 
The experimental results not only report the new SOTA performance achieved by CycleTrans with great advantages to existing methods, \emph{e.g.}, $+4.57\%$ Rank-1 and $+2.2\%$ Rank-1 on SYSU-MM01 and RegDB, but also greatly validate the effectiveness of our method towards the modality gap. 

\section{Acknowledgements}

This work was supported by the National Science Fund for Distinguished Young Scholars (No.62025603), the National Natural Science Foundation of China (No. U21B2037, No. 62176222, No. 62176223, No. 62176226, No. 62072386, No. 62072387, No. 62072389, and No. 62002305), Guangdong Basic and Applied Basic Research Foundation (No. 2019B1515120049), and the Natural Science Foundation of Fujian Province of China (No. 2021J01002).

\bibliography{main}

\end{document}